# Real-Time Robot Localization, Vision, and Speech Recognition on Nvidia Jetson TX1

Jie Tang[1], Yong Ren[2], Shaoshan Liu[2]

[1] South China University of Technology, China
[2] PerceptIn
{cstangjie@scut.edu.cn, yong.ren@perceptin.io, shaoshan.liu@perceptin.io}

**Abstract.** Robotics systems are complex, often consisted of basic services including SLAM for localization and mapping, Convolution Neural Networks for scene understanding, and Speech Recognition for user interaction, etc. Meanwhile, robots are mobile and usually have tight energy constraints, integrating these services onto an embedded platform with around 10 W of power consumption is critical to the proliferation of mobile robots. In this paper, we present a case study on integrating real-time localization, vision, and speech recognition services on a mobile SoC, Nvidia Jetson TX1, within about 10 W of power envelope. In addition, we explore whether offloading some of the services to cloud platform can lead to further energy efficiency while meeting the real-time requirements.

## 1   Introduction

In our view, robots should have three basic capabilities: to localize itself, to understand the environment it sees, and to understand human commands through speech. The core technologies enabling these capabilities are Simultaneous Localization And Mapping (SLAM) [1] and Convolution Neural Networks (CNN) [2], and speech recognition [3, 4].  SLAM is the problem of constructing or updating the map of an unknown environment while simultaneously keeping track of the location of the agent. It is a complex pipeline that consists of many computation-intensive stages, each performing a unique task. CNN consists of multiple processing layers that attempt to model high-level abstractions in the data. In recent years, CNN has achieved significant improvements with computer vision over traditional methods. Speech recognition can either utilize Gaussian Mixture Model (GMM) [3] or Deep Neural Network (DNN) [4] for speech frame classification, and then these two approaches can be used together with the Hidden Markov Model (HMM) and Viterbi algorithm to decode frame sequences.

The rise of these new applications impose two main challenges: first, these workloads have complex pipelines and are computation intensive and they often have tight real-time requirements. For example, in SLAM, sensor data can rush in at a rate as high as 1 KHz, meaning that the computation pipeline needs to process sensor data to produce 1,000 position data in a second, it also means that the longest stage of the pipeline cannot take more than 1 ms to process. In addition, the samples form a time series and are independent. This means that the incoming samples cannot be processed in parallel. For CNN, the camera may similarly capture pictures at a frame rate of 60 FPS, meaning that the CNN pipeline needs to extract meaningful features and recognize the objects within 16 ms. Similarly, speech recognition imposes strong real-time requirements in order for the experience to be interactive. Second, these workloads run on battery-powered mobile robots with extremely limited energy budget. Therefore, it is thus imperative to optimize power consumption in these scenarios as well.

In this paper, we present a case study on integrating real-time localization, CNN-based object recognition, and speech recognition services on a heterogeneous mobile SoC, Nvidia Jetson TX1 [5]. Through this case study, we demonstrate that, with efficient utilization of the heterogeneous computing resources, within about 10 W power consumption, we can simultaneously enable localization, vision, and speech recognition. In addition, we study whether offloading some of the services to cloud can lead to energy efficiency while maintaining the real-time requirements.

## 2 Basic Robotic Services

In this section, we go through the basic robotic services we implement on TX1, including SLAM, CNN-based object recognition, and GMM-HMM-based speech recognition.

### 2.1 SLAM

SLAM is the problem of constructing or updating a map of an unknown environment while simultaneously keeping track of an agent's location within it [1]. Fig. 1 shows a simplified version of the general SLAM pipeline which operates as follows:

1. The *Inertial Measurement Unit*, or IMU, consists of a gyroscope to measure angular velocity and accelerometers to measure acceleration in the three axis. The IMU produces six data points (angular velocities in three different axis and accelerations in the three axis) at a high rate and feeds the data to the propagation stage.

2. The main task of the *Propagation Unit* is to integrate the IMU data points and produce a new position. Since IMU data is received at fixed intervals, by integrating the accelerations twice over time, we can derive the displacement of the agent during the last interval. However, since the IMU hardware usually has bias and inaccuracies, we cannot fully rely on Propagation data, lest the positions produced gradually drift from the actual path.

3. To correct the drift problem, we use a camera to capture frames along the path at a fixed rate, usually at 60 FPS.

4. The frames captured by the camera can be fed to the *Feature Extraction Unit*, which extracts useful corner features and generates a descriptor for each feature.

5. The features extracted can then be fed to the *Mapping Unit* to extend the map as the Agent explores. Note that by map, we mean a collection of 3D points in space, each 3D point would correspond to one or more feature points detected in the *Feature Extraction Unit*.

6. Also, the features detected would be sent to the *Update Unit* which compares the features to the map. If the detected features already exist in the map, the Update unit can then derive the agent's current position from the known map points. By using this new position, the Update Unit can correct the drift introduced by the Propagation Unit. Also, the Update unit updates the map with the newly detected feature points.

In this implementation, we use our proprietary SLAM system that utilizes a stereo camera for image generation at 60 FPS, with each frame having the size of 640 by 480 pixels. In addition, the IMU device generates 200 Hz of IMU updates (three axis of angular velocity and three axis of acceleration).

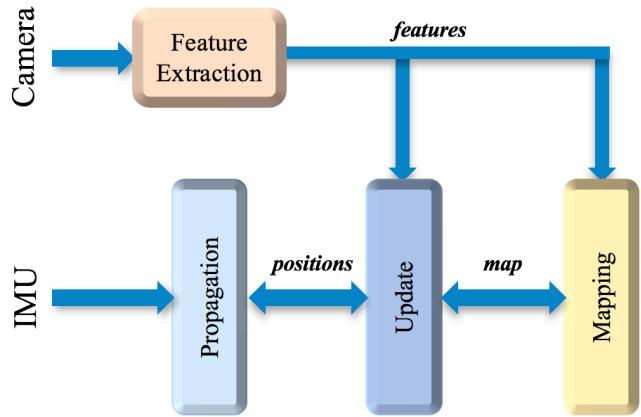

**Fig 1. Visual Inertial SLAM Execution**

### 2.2 Object Recognition with Convolution Neural Network

Convolution Neural Network (CNN) is a type of Deep Neural Network that is widely used in object recognition tasks [2]. Fig. 2 shows a simplified version of the general CNN evaluation pipeline, which usually consists of the following layers:

- The Convolution Layer contains different filters to extract different features from the input image. Each filter contains a set of "learnable" parameters that will be derived after the training stage.
- The Activation Layer decides whether to activate the target neuron or not. Common activation functions include the saturating hyperbolic tangent function, the sigmoid function, and the rectified linear units.

- The Pooling Layer reduces the spatial size of the representation to reduce the number of parameters and consequently the computation in the network.
- The Fully Connected Layer where neurons have full connections to all activations in the previous layer. It derives the labels associated with the input data.

Note that in a normal network we would have multiple copies of the Convolution, Activation, and Pooling Layers. This way, the network can first extract low-level features, and from the low-level features it derives high-level features, and at the end it reaches the Fully Connected Layer to generate the labels associated with the input image. In our implementation, we use the Jetson inference engine provided on Nvidia TX1 platform [9].

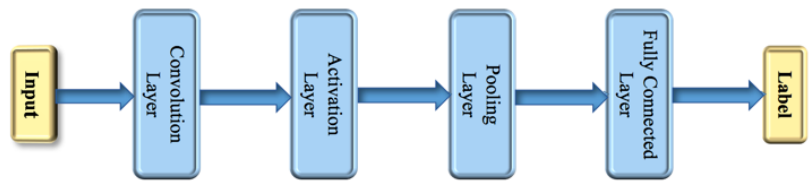

**Fig 2. Simplified CNN Inference Engine**

## 2.3 Speech Recognition

For speech recognition we choose GMM over DNN for classification since GMM is faster to compute and GMM models are easier to derive. Although DNNs are more accurate classifiers, they are very slow to train compared to GMM and they usually impose high computing power requirements, and may not be suitable for mobile devices. A generic pipeline of speech recognition is shown in Figure 3, which can be divided into the following stages:
:
- First the speech signal goes through the feature extraction stage, which extracts feature vector. In this case, we utilize a GMM-based feature extractor.
- Then the extracted feature vector is fed to the decoder, which takes an acoustic model, a pronunciation dictionary, and a language model as input. The decoder then decodes the feature vector into a list of words.

Note that in this implementation we utilized the speech model presented in [12], which uses GMM for classification and HMM for decoding.

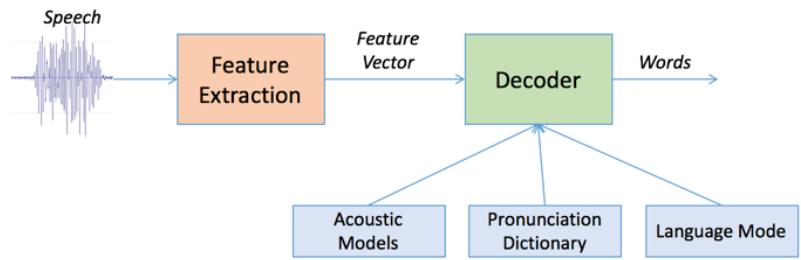

Fig 3. Speech Recognition Engine

## 3  Robotic System on Jetson TX1

In this section we present how we implement the aforementioned robotic services onto a Jetson TX1 SoC to maneuver a mobile robot chassis. We will go through the hardware setup, the system architecture, as well as performance and power consumption of such implementation.

### 3.1  Hardware Setup

To implement the services, we first need a hardware setup for the robot. The robot consists of two parts: the perception and decision units, which are implemented on TX1, and the execution unit, which is the robot chassis. The robot chassis receives commands from the TX1, and executes these commands accordingly.

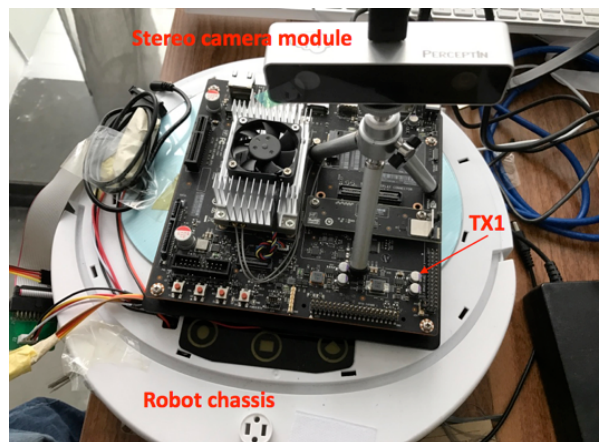

Fig 4. Robot setup

Figure 4 shows the hardware setup of the implementation. First, a stereo visual inertial camera module [11] is connected to the TX1 board through USB 3.0 interface.

This module generates VGA-resolution stereo images at 60 FPS along with IMU updates at 200 Hz. These raw data are fed to the SLAM pipeline to produce accurate location updates, and fed to the CNN pipeline to perform object recognition. Also, the TX1 board is connected to the underlying chassis through a serial connection. This way, after going through the perception and decision stages, TX1 sends commands to the underlying chassis to move it around. For instance, after the SLAM pipeline produces a map of the environment, the decision pipeline can instruct the robot to move from location A to location B, and the commands are sent through the serial interface. For speech recognition, to emulate commands, we set a thread to constantly perform audio playback to the speech recognition pipeline. In addition, a 2200 mAh battery is used to power the TX1 board.

### 3.2 System Architecture

Once we have made a decision on the hardware setup, the next challenge is to design a system architecture to tightly integrate these services. Figure 5 presents the architecture of the system we implement on TX1. At the frontend, we have three sensor threads to generate raw data: the camera thread generates images at a rate as high as 60 Hz, the IMU thread generates inertial updates at a rate of 200 Hz, and the microphone thread generates audio signal at a rate of 8 KHz. The image and IMU data then get into the SLAM pipeline to produce a position update at a rate of 200 Hz. Meanwhile, as the robot moves, the SLAM pipeline also extends the environment map. The position updates, along with the updated map, then get passed to the navigation thread to decide how the robot makes its next move. The image data also gets into the object recognition pipeline to extract the labels of the objects that the robot encounters. The labels of the objects then get fed into the reaction unit, which contains a set of rules of what to do next when a specific label is detected. For instance, a rule can be that whenever a human face is detected, the robot should greet the person. The audio data gets through the speech recognition pipeline to extract commands, and then commands are fed to the command unit. The command unit stores a set of predefined commands, and if the incoming command matches one in the predefined command interface, then an action is triggered. For instance, we implement a command "stop", whenever the word "stop" is heard, the robot stops all its ongoing actions.

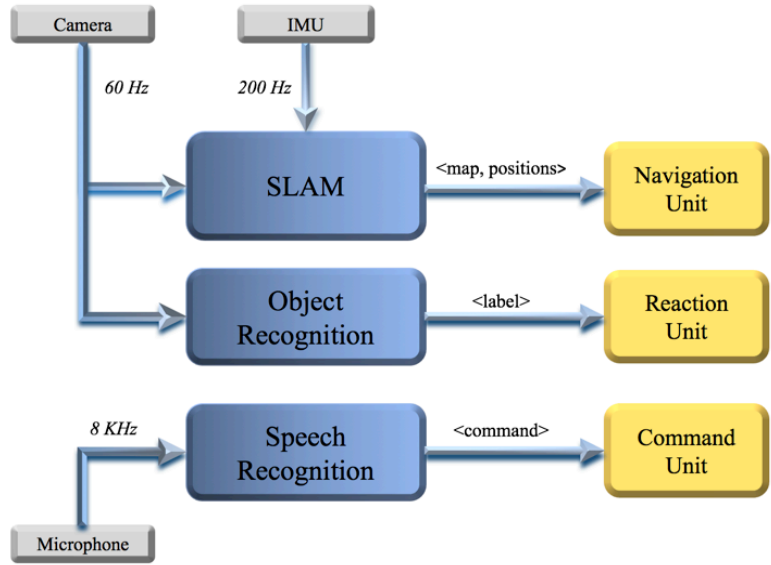

**Fig 5. System Integration**

This architecture provides very good separation of different tasks, with each task hosted in its own process. The key to high performance and energy efficiency is to fully utilize the underlying heterogeneous computing resources for different tasks. For instance, feature extraction operations used in the frontend of SLAM as well as CNN computations exhibit very good data parallelism, thus it might be beneficial to offload these tasks to GPU, and free up some CPU resources for other computation, or for energy efficiency. Therefore, in our implementation, the SLAM frontend is offloaded to GPU, the SLAM backend is executed on CPU; the major part of object recognition is offloaded to GPU; the speech recognition task is executed on CPU. We will explore how this setup behaves on the Jetson TX1 SoC in the next subsections.

### 3.3 Performance Evaluation

In this subsection we study the performance of this system. When running all the services on the system, the SLAM pipeline can consume images at 10 FPS on TX1 if we use CPU only. However, once we accelerate the feature extraction stage on GPU, the SLAM pipeline can consume images at 18 FPS, almost doubles the performance of the CPU-only version. In our practical experience, once the SLAM pipeline is able to consume images at more than 15 FPS, then we have a stable localization services. Also as a reference, we also measured the SLAM performance on an Intel Core i5 CPU, at its peak, the SLAM pipeline consumes images at 15 FPS. Therefore, with the help of GPU, the TX1 SoC can outperform Intel CPU on SLAM workloads. For the vision deep learning task (using Jetson Inference engine), can process 30 FPS in image recognition. This task is mostly GPU-bound. For our robotic application, we don't actually need 30 FPS frame rate for object recognition as the robot travels at fairly low speed (at 1 m/s), a 10 FPS should satisfy our needs. For the speech recognition, which we use Kaldi and it is CPU-bound, we can convert an audio stream into words within 100 ms latency. In our requirement, we can tolerate 500 ms latency in such tasks. In summary, to our surprise, after we enable all these services on, TX1 can still satisfy the real-time performance requirement. The main reason is that GPU performs most of the heavy lifting, especially on SLAM and vision tasks.

Next we are curios about the system resource utilization when running these tasks. As shown in Figure 6, when running the SLAM task, it consumes about 28% CPU, 2% GPU, and 4% of system memory. The GPU is mainly used to accelerate feature extraction in this task. When running speech recognition, it consumes about 22% CPU, no GPU, and 2% system memory. For vison-based deep learning task, it consumes 24% CPU, 70% GPU, and 22% of system memory. When combining all three tasks together, the system consumes 60% CPU, 72% GPU, and 28% of system memory, still leaving enough headroom for other tasks.

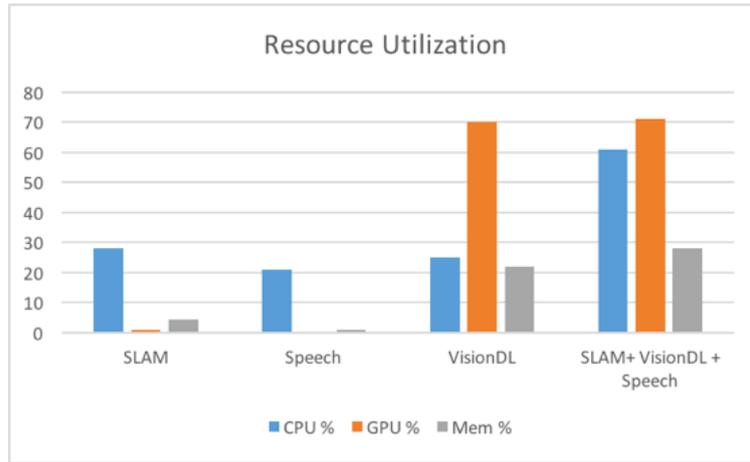

**Fig 6. Resource utilization on TX1**

Next we would like to understand the power consumption behavior. As shown in Figure 7, even when running all these tasks simultaneously, the TX1 module only consumes 11 W, the GPU consumes 3.5 W, and the CPU consumes 4.2 W. In other words, within a 11 W power envelope, we can enable real-time robot localization, object recognition, and speech recognition on a TX1 SoC module.

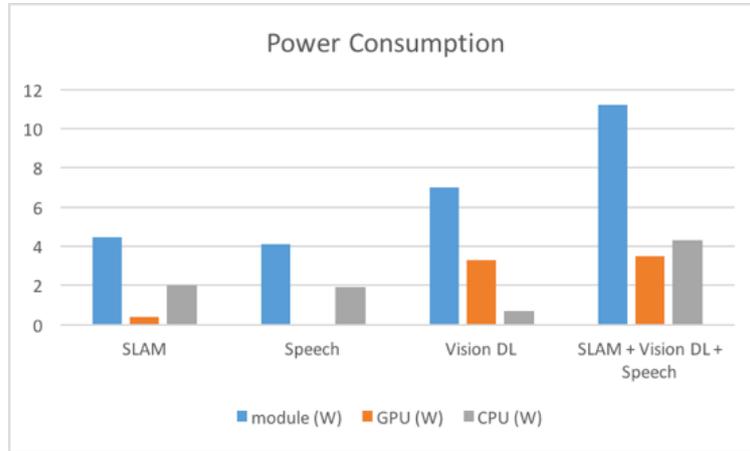
**Fig 7. Power consumption on TX1**

## 4 Offloading Cloud vs. Computing Locally

In the previous section, we learn that we can enable real-time robot localization, vision, and speech tasks on TX1 within 11 W power envelope. A 2200 mAh battery can power the device for about two hours. In this section, we try to explore whether offloading some of the robotic workloads to the cloud can be a viable solution to reduce energy consumption while meeting real-time requirements. Out of the three services, SLAM is not a good candidate for offloading since the robot requires a position update every 5 ms, any delay will result in incorrect localization behavior. On the other hand, for objection recognition and speech recognition, we can tolerate > 100 ms latency, making them potential candidates for offloading.

First, we try to offload the object and speech recognition tasks to a remote server, outside of the local area network. The problem with this approach is that the latency is high for vision and speech recognition tasks, ranging from 2 to 5 seconds, which does not satisfy the real-time requirement. Next, we deploy the cloud within the local area network. After making this configuration, we have a local cloud that can perform object recognition within 100 ms and speech recognition within 200 ms, meeting the real-time requirement for the robot deployment.

Regarding resource utilization and power consumption, figure 8 shows the resource utilization of offloading the services compared to executing locally. When offloading the tasks, we send the image or the audio file to the cloud and then wait for the resource. For speech recognition, offloading consumes 5% of CPU vs. 20% CPU when executing locally. For object recognition, offloading consumes 12% CPU vs. 25% CPU and 70% GPU. When offloading object and speech recognition tasks and executing the SLAM task locally, the module consumes 5 W, and under this configuration a 2200 mAh battery can power the device for about five hours.

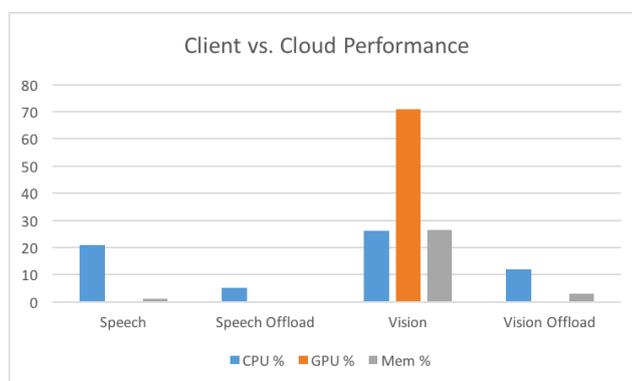
**Fig 8. Client vs Cloud Performance**

Based on the results from this section, we conclude that to maximize mobile robot battery life, we can deploy local clouds to host object recognition and speech recognition tasks. Especially in a multiple robot deployment, a local cloud can be shared by the robots.

## 5   Related Work

To our knowledge, this is the first work that integrates robotic localization, vision, and speech capabilities onto an embedded SoC with about 10 W of power consumption. Nonetheless, many previous implementations have integrated multiple services onto one robot. ASIMO is a humanoid robot implemented by Honda [13] which integrates computer vision, speech recognition, as well as mechanical control capabilities onto the same robot. The implementation uses more than twenty processors whereas in our implementation we implement all services on one SoC and the computing system consumes over 100 W. Baxter is another robot that integrates multiple services [14], and it uses an Intel Core i7 CPU as the main processor, at its peak, it consumes around 100 W. Autonomous vehicles are of one extreme form of robots that incorporate localization and computer vision services. In one implementation [15], the autonomous driving computing system utilizes multiple GPU cards along with a high-end Intel Core i7. At its peak, it consumes more than 3000 W of power.

## 6   Conclusions

Integrating multiple services onto one robot system is a complex mission, in order to meet the heavy computing requirements, most existing implementations utilize high-end CPUs which consume more than 100 W. Nonetheless, robots are mobile systems with strict energy constraints. In order for mobile robots to proliferate we need to simultaneously enable all these services within about 10 W of power budget while meeting the real-time requirements. In this paper, we present our experiences of integrating localization, vision, and speech recognition services onto a mobile robot. By carefully designing the system architecture, we manage to simultaneously enable all these services onto Nvidia Jetson TX1 SoC with about 10 W of power consumption. They key to achieve this is to efficiently utilize the heterogeneous computing resources provided by the SoC, using GPU mostly for the frontend of SLAM tasks and for convolution neural network workloads, and using CPU for other tasks. To achieve further energy efficiency, we try to offload computer vision and speech recognition tasks to the cloud. Our study shows that if the cloud is deployed within local area network, by offloading these tasks we can easily double the battery life of the robot while still meeting the real-time requirements.

## References

[1] S. Thrun and J.J. Leonard, Simultaneous Localization and Mapping, Springer Handbook of Robotics, Springer, 2008.

[2] A. Krizhevsky, I. Sutskever, and G.E. Hinton, Imagenet classification with deep convolutional neural networks. In Advances in neural information processing systems (pp. 1097-1105).

[3] D.A. Reynolds, T.F. Quatieri, and R.B. Dunn, Speaker verification using adapted Gaussian mixture models. Digital signal processing, 10(1-3), pp.19-41.

[4] G. Hinton, L. Deng, D. Yu, G.E. *et al*. Deep neural networks for acoustic modeling in speech recognition: The shared views of four research groups. IEEE Signal Processing Magazine, 29(6), pp.82-97.

[5] Nvidia Jetson TX1: https://developer.nvidia.com/embedded/buy/jetson-tx1

[6] O.J. Woodman, An introduction to inertial navigation, University of Cambridge Technical Report, UCAM-CL-TR-696, 2007

[7] E. Jones and S. Soatto, "Visual-inertial navigation, mapping and localization: A scalable real-time causal approach," International Journal of Robotics Research, vol. 30, no. 4, Apr. 2011.

[8] J. Kelly and G. Sukhatme, "Visual-inertial sensor fusion: Localization, mapping and sensor-to-sensor self- calibration," International Journal of Robotics Research, vol. 30, no. 1, Jan. 2011.

[9] Jetson Inference: https://github.com/dusty-nv/jetson-inference

[10] D. Povey, A. Ghoshal, G. Boulianne *et al*. The Kaldi speech recognition toolkit. In IEEE 2011 workshop on automatic speech recognition and understanding (No. EPFL-CONF-192584). IEEE Signal Processing Society.

[11] PerceptIn Stereo Visual Inertial Camera Module: https://www.perceptin.io/product

[12] M. Korvas, O. Plátek, O. Dusek, I. Zilka, and F. Jurcícek, Free English and Czech telephone speech corpus shared under the CC-BY-SA 3.0 license. In LREC (pp. 4423-4428).

[13] M. Hirose, K. Ogawa, Honda humanoid robots development. Philosophical Transactions of the Royal Society of London A: Mathematical, Physical and Engineering Sciences, 365(1850), pp.11-19.

[14] Baxter Robots: http://www.rethinkrobotics.com/baxter/tech-specs/

[15] S. Liu, J. Tang, Z. Zhang, and J.L. Gaudiot, CAAD: Computer Architecture for Autonomous Driving. arXiv preprint arXiv:1702.01894.